\documentclass{bmvc2k}
\usepackage{times}
\usepackage{epsfig}
\usepackage{graphicx}
\usepackage{amsmath}
\usepackage{amssymb}
\usepackage{xcolor}
\usepackage{multirow}

\usepackage{times}
\usepackage{epsfig}
\usepackage{graphicx}
\usepackage{amsmath}
\usepackage{amssymb}
\usepackage{xcolor}
\usepackage{multirow}
\usepackage{subfigure}

\title{STEP CATFormer: Spatial-Temporal Effective Body-Part Cross Attention Transformer for Skeleton-based Action Recognition}

\addauthor{Bao Long Nguyen Huu}{tp22010-0251@sti.chubu.ac.jp}{1}
\addauthor{Tohgoroh Matsui}{matsui@isc.chubu.ac.jp}{1}

\addinstitution{
 Department of Information Engineering \\
 Chubu University\\
 Kasugai, Aichi, Japan
}

\runninghead{Student, Prof, Collaborator}{BMVC Author Guidelines}

\begin{document}

\maketitle

\begin{abstract}
Graph convolutional networks (GCNs) have been widely used and achieved remarkable results in skeleton-based action recognition.
We think the key to skeleton-based action recognition is a skeleton hanging in frames, so we focus on how the Graph Convolutional Convolution networks learn different topologies and effectively aggregate joint features in the global temporal and local temporal. In this work, we propose three Channel-wise Tolopogy Graph Convolution based on Channel-wise Topology Refinement Graph Convolution (CTR-GCN). Combining CTR-GCN with two joint cross-attention modules can capture the upper-lower body part and hand-foot relationship skeleton features.
After that, to capture features of human skeletons changing in frames we design the Temporal Attention Transformers to extract skeletons effectively.
The Temporal Attention Transformers can learn the temporal features of human skeleton sequences. Finally, we fuse the temporal features output scale with MLP and classification. We develop a powerful graph convolutional network named Spatial Temporal Effective Body-part Cross Attention Transformer which notably high-performance on the NTU RGB+D, NTU RGB+D 120 datasets. Our code and models are available at \href{https://github.com/maclong01/STEP-CATFormer}{https://github.com/maclong01/STEP-CATFormer}
\end{abstract}


\begin{figure*}
\begin{center}
\includegraphics[width=0.9\linewidth]{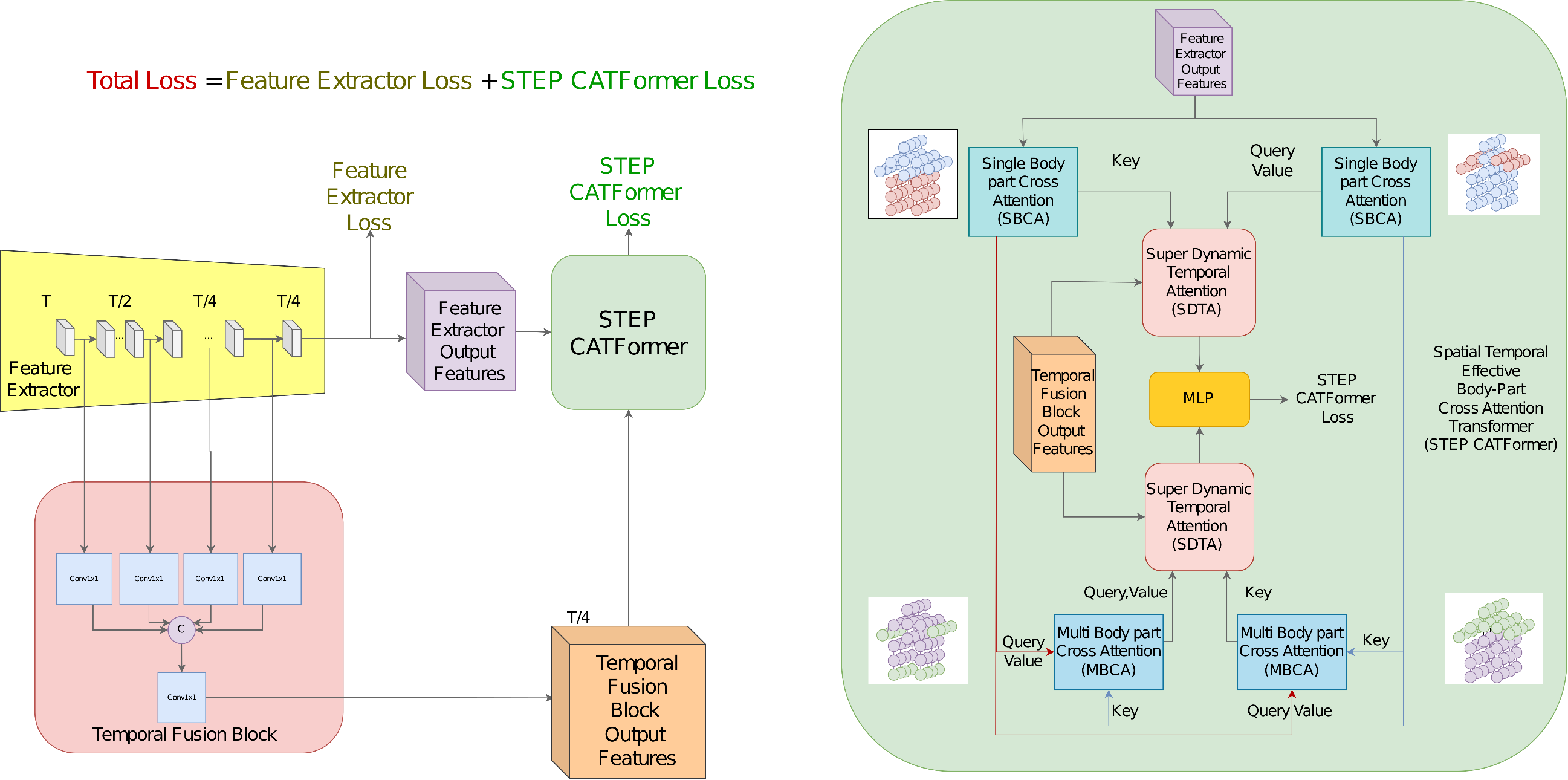}
\end{center}
   \caption{\textbf{STEP CAT model } has 3 branches Feature extractor branch, temporal fusion branch, and STEP CATFormer branch. In STEP CATFormer branch has 4 small branches Single Body-part Cross Attention (\textbf{SBCA}), Multi Body-part Cross Attention (\textbf{MBCA}), Super Dynamic Temporal Attention (\textbf{SDTA}), and MLP with classification.}
\label{fig-stepcat}
\end{figure*}

\section{Introduction}
Computer vision is a field that has widespread applications in various aspects of life, such as object recognition, image segmentation, and human action recognition. In recent years, human action recognition has received significant attention due to advancements in deep learning and computer vision. Applications of human action recognition include games play, eldercare, healthcare assistance, and video surveillance. With the development of high-performance sensors and advanced algorithms for human pose estimation \cite{Fang_2017_ICCV}, it is now possible to acquire accurate 3D skeletal data. Recent advances in 3D depth cameras such as Microsoft Kinect camera \cite{Kinect_2017_ICCV} was an attempt to broaden the 3D gaming experience of the Xbox 360's audience and advanced human pose estimation algorithms \cite{Cao_2019_IEEE} enable quick and accurate adjustment of 3D skeletons using inexpensive devices.
In computer vision for games, Kinect sensor real-time skeletal tracking recognizes the actions of the human body is the key technology behind Kinect is humanbody language understanding, which means that the computer first recognizes and understands what the user is doing, before responding exactly what they need to improve the quality of the game play. But skeleton-based action recognition still faces several challenges, including body size, viewpoint, and motion speed \cite{Yan_2018_AAAI,Aggarwal_2011,Ren2020-eh}.

However, Graph Neural Networks (GNNs) \cite{graphnn2018}, and specifically, Graph Convolutional Networks (GCNs)\cite{SGCN2019}, can effectively capture spatial and temporal information to solve various problems. Yan et al. \cite{Yan_2018_AAAI} first applied GCNs to the field of skeleton-based action recognition and showed that the joints of the skeletal spatiotemporal graph of human action are correlated. They used GCNs and temporal convolution \cite{temporalconv2018} to extract motion features and graph structures to model the correlation between human joints in space-time. Various approaches to skeleton-based human action recognition have been proposed based on this idea, including the use of second-order information from skeletal data, multi-stream networks, and attentional mechanisms \cite{Shi2018-nh,Cheng_2020_CVPR,Qin2021-wu,Plizzari2020-lc,Zhang2019-oa}.

However, these approaches have limitations such as limiting the representational power of the model in channel topologies, having unnaturally connected joint relationships, and ignoring the variability of different channel topologies. To address these limitations, Chen et al. \cite{Chen2021-lx} proposed Channel Topology Refinement Graph Convolution (CTR-GCN), an approach that learns topologies and aggregates features in different channel dimensions dynamically and now many method \cite{Cheng_2020_CVPR,Fanfan_2020,Zhan_Chen2022} base on it. However, it mostly favors modeling in the spatial dimension and does not emphasize temporal dimensions. 

To improve temporal features and consider the features importance in different body parts and joints, we propose a method called Spatial Temporal Effective Body-part Cross Attention Transformer. This method can dynamically learn the relationship between body parts and joints in body parts features on spatiotemporal dimensions. Specifically, we use CTR-GCN's dynamic channels topologies temporal feature representation in high-dimensional space extracted in the last layer to capture features in body parts with intra-body joints in body parts relationships. We also propose a powerful temporal attention mechanism to efficiently extract temporal features using low-dimensional space extracted in the previous layer of CTR-GCN and fuse them with embedding body parts features. The body parts and intra-body part attention method combined with the powerful temporal attention transformers completes the modeling.




\section{Related work}

With the rise of vision transformers, transformer-based methods have been applied to skeleton data analysis, such as \cite{Plizzari2020-lc,Shi2020-uk,Wang2021-ua,Kim2022-pu,Gao2022-pd,Wangmeng-2022}. Recent studies have extended the Transformer model to the recognition of actions based on skeleton data in both spatial and temporal dimensions \cite{Plizzari2020-lc,Shi2020-uk,Wang2021-ua}. The IIP-Transformer model \cite{Wang2021-ua} was the first to use self-attention to understand the relationships between joints, while some datasets employ a combination of spatial transformer and temporal transformer. LST \cite{Wangmeng-2022} uses a hybrid architecture that combines GCN and Transformer in a body part, where each body-part have using contrastive learning before spatial in place of GCN and temporal convolution. These methods effectively capture spatiotemporal information about the skeleton and show promise in skeletal action recognition. Our work in this paper we combine graph convolution and transformer method hybrid architecture model training for skeleton-based action recognition.


\section{Proposed Method}

\subsection{Spatial-Temporal Effective Body-part Cross Attention Transformer Overview}
An overview of the proposed STEP-CATFormer network is shown in Figure \ref{fig-stepcat}. It consists of three types of blocks (Feature Extractor, Temporal Fusion, and Spatial-Temporal Effective Body-part Cross Attention Transformer), in STEP-CATFormers which our three primary components with spatial dimension are Single Body-part Cross Attention Block and Multi Body-part Cross Attention Block, and Temporal Dimension Super Dynamic Temporal Attention. STEP-CATFormers focus on modeling both the discriminative relationships between joints and body part skeletons in spatio-temporal motion patterns for recognition. First, given a skeleton sequence from the feature extractor $X_{in}\in\mathbb{R}^{N\times T\times C_{0}}$, a linear layer is applied to the STEP-CATFormer to project it to the position embedding, generating the feature $X_{1}\in\mathbb{R}^{N\times T\times C_{1}}$.
Then, $X_{1}$ is split and passed into two Single Body-part Cross Attention (SBCA) branches. One branch adaptively discriminates the relationship in the hand joints and another joint, producing features $\{X^{H}_{2}\in\mathbb{R}^{H\times T\times C}\}$, where  $H$ is the number of hand joints. And, the other branch partitions discriminate the relationship in the leg-foot joints and another joint, generating the feature $\mathrm{X}^{F}_{2}\in\mathbb{R}^{F\times T\times C}$, where $F$ is the number of foot and leg joints. $X^{H}_{2}$ and $X^{F}_{2}$ are then passed through Super Dynamic Temporal Attention (SDTA) and Multi Body-part Cross Attention (MBCA). In one of the ways, the inputs $X^{H}_{2}$ and $X^{F}_{2}$ are passed into two MBCA branches. One branch adaptively discriminates the relationship in the wrist-ankle joints and another joint part, producing features $X^{\mathit{WA}}_{3}\in\mathbb{R}^{\mathit{WA}\times T\times C}$, where $\mathit{WA}$ is the number of wrist-ankle joints. Other branch partitions discriminate the relationship in the up and down joints and another joint generates the feature $\mathrm{X}^{\mathit{UD}}_{3}\in\mathbb{R}^{\mathit{UD}\times T\times C}$, where $\mathit{UD}$ is the number of up and down joints. In particular, the SBCA, MBCA block consists of a spatial transformer sub-block and a joint-part cross-attention sub-block to sufficiently model the spatial interaction information of actions. In the Super Dynamic Temporal Attention Transformer with the input $X^{H}_{2}$ and $X^{F}_{2}$ and temporal fusion features to give output $\{X^{T_{1}}_{n}\in\mathbb{R}^{T\times C}\}^{N}_{n=1}$. For other input features $X^{\mathit{WA}}_{3}$ and $X^{\mathit{UD}}_{3}$, the SDTA gives the output features at $\{X^{T_{2}}_{n}\in \mathbb{R}^{T\times C}\}^{N}_{n=1}$. Finally, using two outputs of the SDTA to do element-wise addition, then processing out with MLP to produce features and applied Global Average Pooling (GAP) $X_{out}\in\mathbb{R}^{1\times 1\times C_{out}}$ to classification using fully connected layer and Softmax classifier.

\subsection{Single Body-part Cross Attention}

\begin{figure}[tbp]
  \begin{minipage}[b]{0.48\columnwidth}
    \centering
    \includegraphics[width=\columnwidth]{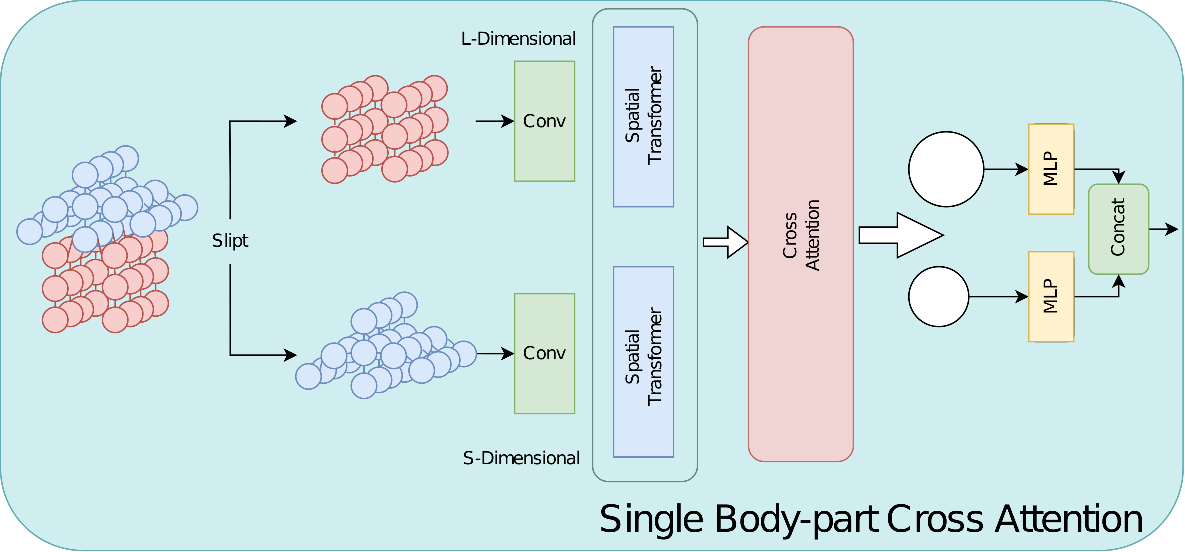}
    \label{fig-sbca}
    \caption{\textbf{Single Body-part Cross Attention} (SBCA) had input features from feature extractor to two Spatial Attention with one L Transformer proposed Large channels dimension Transformer and S Transformer 
    }
  \end{minipage}
  \hspace{0.02\columnwidth} 
  \begin{minipage}[b]{0.48\columnwidth}
    \centering
    \includegraphics[width=\columnwidth]{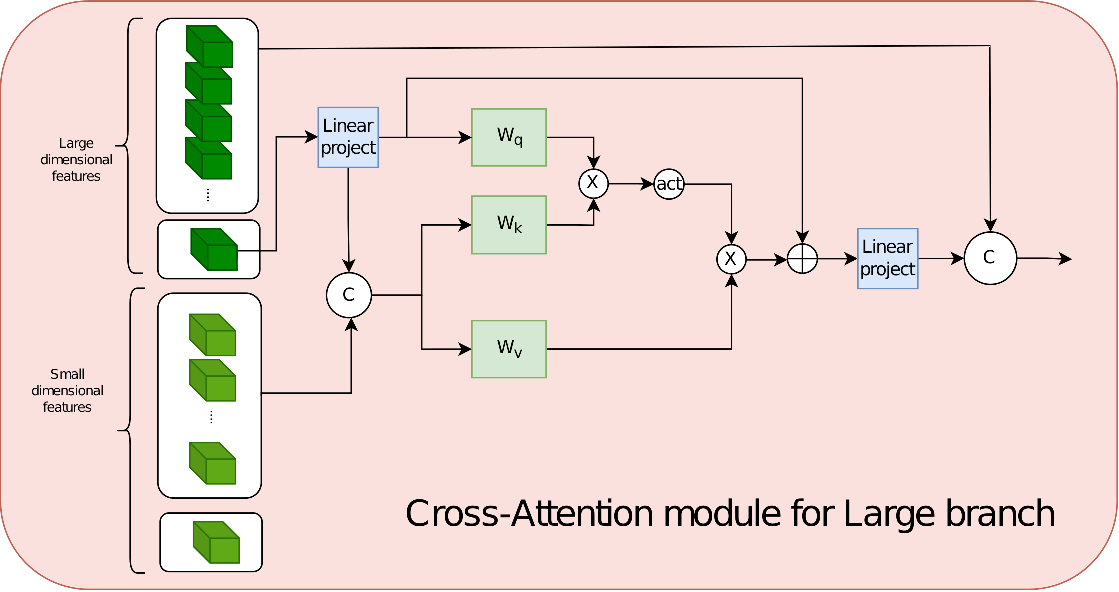}
    \label{fig-cross}
    \caption{\textbf{Cross-attention module for Large branch}. The joint tokens of the large-dimension transformer serve as a query token to interact with the joint tokens from the small-dimension transformer. Linear Projects are projections to align dimensions.
    }
  \end{minipage}
\end{figure}

To allow information to diffuse across the joints and body parts, we develop a Single Body-part Cross Attention (SBCA) block with a cross-attention sub-block, shown in Figure \ref{fig-sbca}. Cross-attention uses multi-head cross-attention to interact and diffuse the features of the two branches. We describe the cross-attention module for the large branch (L-branch) with the input as $\{x^{F}_{2}\in\mathbb{R}^{F\times T\times C}\}$, where $F$ is the number of leg and foot joints extracted by $1\times1$ convolution and spatial transformation in large dimension, and the same procedure is performed for the small branch (S-branch) with the input as $\{x^{O}_{2}\in\mathbb{R}^{O\times T\times C}\}$, where $O$ is the number of the remaining joints, excluding the foot and leg joints extracted by $1\times1$ convolution and spatial transformer in small dimension and by simply swapping the index $l$ and $s$. The cross-attention module for the large branch is shown in Figure \ref{fig-cross}. For branch $l$, it first collects the tokens from the S-Branch and concatenates them, as shown in
\begin{equation}
{(x^{F})'}^{l}_{cls}=f^{l}(x^{F})^{l}_{cls},\quad
{(x^{F})'}^{l}=[{(x^{F})'}^{l}_{cls}\Vert \mathrm{(x^{O})}^{s}_{d^{s} - cls}] \text{,}
\end{equation}
where $f^{l}(\cdot)$ is the projection function for dimension alignment and $d^{s}$ is the $s$-branch dimension of $x^{O}$, $\{{(x^{F})'}^{l}_{cls}\in\mathbb{R}^{F\times T\times 1}\}$, and $\{{(x^{O})}^{s}_{d^{s}-cls}\in\mathbb{R}^{F\times T\times C-1}\}$.
Mathematically, the cross-attention can be expressed as
\begin{equation}
q={(x^{F})'}^{l}_{cls}W_{q},\,k={(x^{F})'}^{l}W_{k},\,v={(x^{F})'}^{l}W_{v},\,
A=\mathrm{softmax}(qk^{T}/\sqrt{C/h}),CA((x^{F})'^{l})=Av \text{,}
\end{equation}
where $W_{q},\, W_{k},\, W_{v}\in \mathbb{R}^{C\times(C/h)}$ are the learnable parameters, and $C$ and $h$ are the embedding dimensions and the number of heads, respectively.
Since we use $cls$ in the channel queries, the generation of the attention map A in cross-attention is linear as in all-attention. In self-attention, we also use multiple heads in the cross-attention and represent it as MCA. The output $(y^{F})^{l}_{2}$ of the branch cross-attention module with a large channel dimension is defined as follows:
\begin{align}
(y^{F})^{l}_{cls}&=f^{l}((x^{F})^{l}_{cls})+\mathit{MCA}(\mathit{LN}([f^{l}{(x^{F})}^{l}_{cls}\Vert (x^{O})^{s}_{d^{s} - cls}]) \\
(y^{F})^{l}_{2}&=[g^{l}((y^{F})^{l}_{cls})\Vert (x^{F})^{l}_{d^{l} - cls}] \text{,}
\end{align}
where $\{{(x^{F})}^{l}_{d^{l} - cls}\in\mathbb{R}^{F\times T\times C-1}\}$, and $f^{l}(\cdot)$ and $g^{l}(\cdot)$ are the projection and back-projection functions for dimensionaln alignment, respectively. Finally, after MLP alignment and concatenation, $FFN$ contains a two-layer multilayer perceptron with expansion ratio $r$ at the hidden layer and a GELU non-linearity is applied after the first linear layer. The layer normalization ($LN$) is applied before every block, and residual shortcuts after every block can be expressed as
\begin{equation}
z^{F}_{2}=f_{l}[(y^{F})^{l}_{2}] \Vert f_{s}[({y}^{O})^{s}_{2}],\quad
{X}^{F}_{2}={z}^{F}_{2}+\mathit{FFN}({z}^{F}_{2}) \text{,}
\end{equation}
where $(y^{O})^{s}_{2}$ is the output from the cross-attention small branch, $f_{l}$ and $f_{s}$ are the MLP alignments for the small and large channel dimensions.
The output is ${X}^{F}_{2}$ and the out of ${X}^{H}_{2}$ has the same single body-part cross-attention construction module with ${X}^{F}_{2}$ with non-shared parameters.

\subsection{Multi Body-part Cross Attention}

\begin{figure}[t]
\begin{center}
   \includegraphics[width=0.6\linewidth]{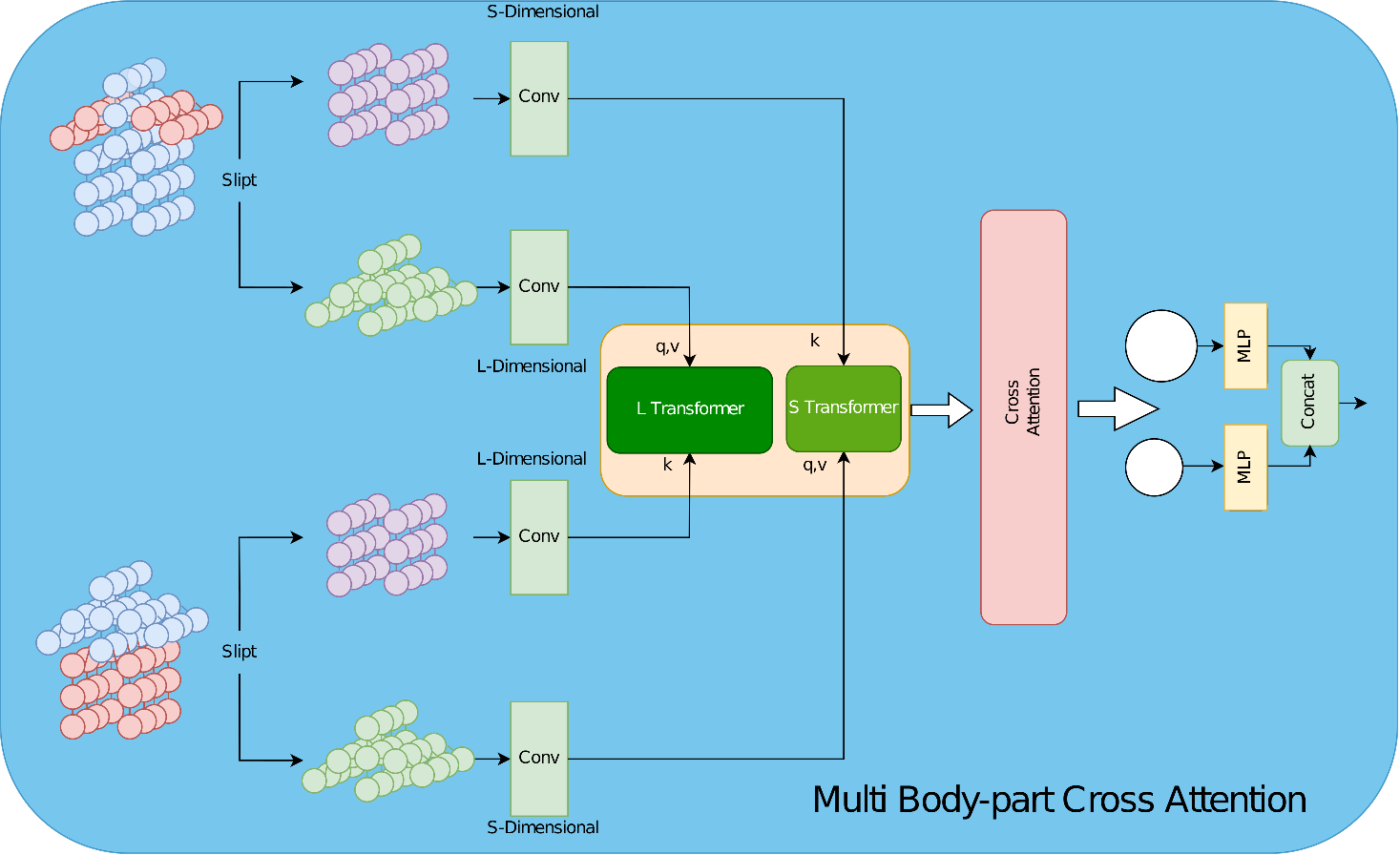}
\end{center}
   \caption{\textbf{Multi Body-part Cross Attention} (MBCA) had input with different attention body-part features to two Spatial Attention with one L Transformer proposed Large channels dimension Transformer and S Transformer represented Small channels dimension Transformer then with two input we use the cross attention method to give the relationship in L and S Transformer features output.}
\label{fig-mbca}
\end{figure}

Second, to defuse more detaild information across the joints and body parts based on the features extracted by SBCA, we develop a Multi Body-part Cross Attention (MBCA) block with cross-attention sub-block, shown in Figure \ref{fig-mbca}.

Like SBCA, the cross-attention uses multi-head cross-attention to interact and diffuse features of the two branches. The cross-attention module for the large branch (L-branch) with the input as $\{{(x)^{l}}^{\mathit{UD}}_{3}\in\mathbb{R}^{\mathit{UD}\times T\times C}\}$, where $\mathit{UD}$ is the number of up and down joints extracted by 2 $1\times1$ convolution and spatial transformer block with 2 features input. One of them is  extracted by hand joints cross-attention block $\{{x}^{U}_{3}\in\mathbb{R}^{U\times T\times C}\}$, where $\mathit{U}$ is the number of up joints that comes to $q$, $v$ gated in spatial transformer by large dimensional convolution. And the other one as extracted in foot-leg joint cross-attention block $\{{x}^{D}_{3}\in\mathbb{R}^{D\times T\times C}\}$, where $D$ is the number of down joints that comes to $k$ gated in spatial transformer by large dimensional convolution. Both the hand joint cross-attention block and the foot-leg joint cross-attention block module constructed by the SBCA module we introduced before.

And the same procedure is performed for the small branch (S-branch) with the input as $\{{x^{s}}^{\mathit{UD}}_{2}\in\mathbb{R}^{\mathit{UD}\times T\times C}\}$.
One of them is extracted by hand joint cross-attention block $\{{x}^{U}_{3}\in\mathbb{R}^{U\times T\times C}\}$, where $U$ is the number of up joints comes to $k$ gated in spatial transformer by small dimensional convolution. And the other one is extracted in foot-leg joint cross-attention block $\{{x}^{D}_{3}\in\mathbb{R}^{D\times T\times C}\}$, where $D$ is the number of down joints come to $q$, $v$ gated in spatial transformer by small dimensional convolution. Similar to the SBCA, the cross-attention module for the large branch is shown in Figure \ref{fig-cross} with different alignment convolution and spatial transformer construction. Finally, the output is the ${X}^{\mathit{UD}}_{3}$ and the out of ${X}^{\mathit{WA}}_{3}$ has the same multi-body-part cross-attention construction module with ${X}^{\mathit{UD}}_{3}$.

\subsection{Super Dynamic Temporal Attention Transformer}
\begin{figure*}
\begin{center}
\includegraphics[width=1.0\linewidth]{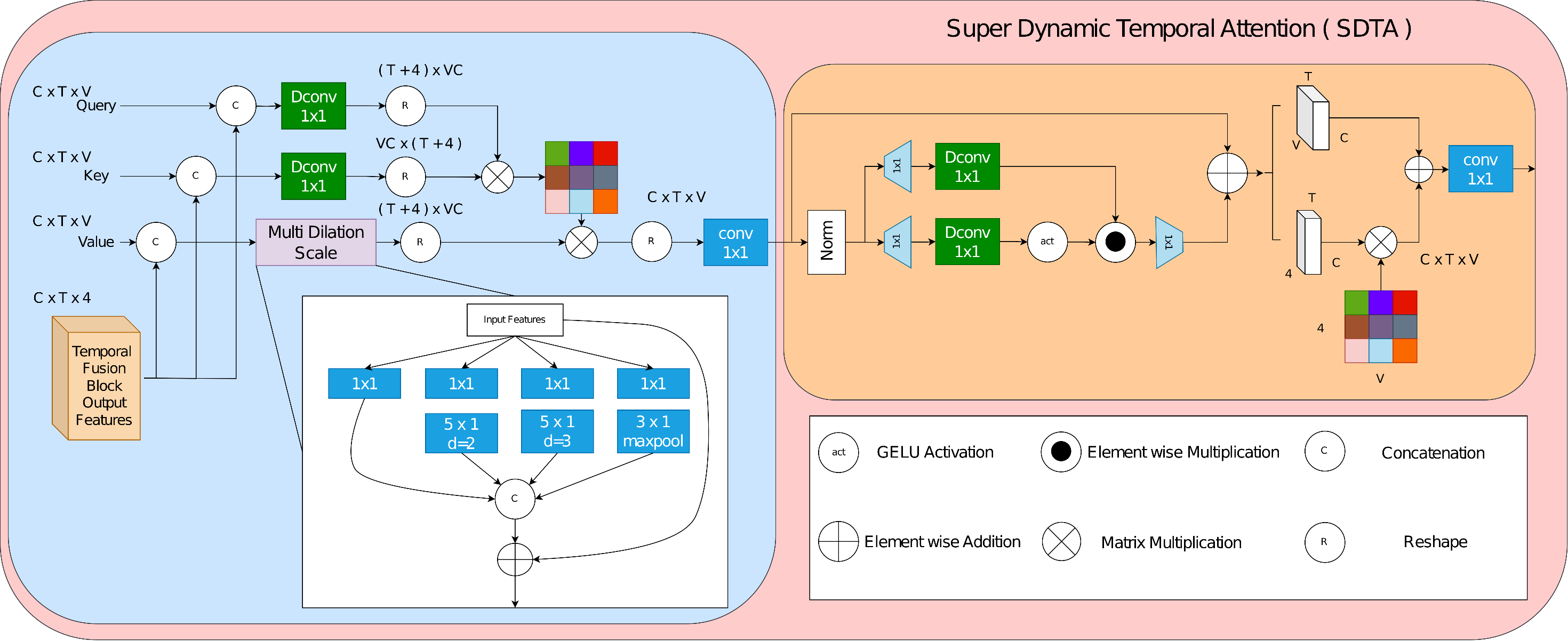}
\end{center}
   \caption{\textbf{Super Dynamic Temporal Attention} (SDTA) has blue area MHA is proposed Multi Head Attention and orange area represented FFN(Feed-Forward Network) have designed base in Gated-Dconv FFN\cite{Syed-2022} and Dynamic Temporal GCN method \cite{Haodong-2022}}
\label{fig-sdta}
\end{figure*}


From a layer normalized tensor $Y \in \mathbb{R}^{N\times T\times C}$, our MHA first generates query $(Q)$, key $(K)$, and value $(V)$ projections, and we perform average pooling to $V$ temporal fusion features ${{X_{1},..., X_{V}|X_{i}\in \mathbb{R}^{C \times T}}}$ to obtain the temporal fusion skeleton-level feature $P$.
Then we used it $P$ concatenation with $\widehat{Q}= Q\Vert P$; $\widehat{K}= K\Vert P$; $\widehat{V}= V\Vert P$ to capture the temporal fusion skeleton features together in the attention method.
This is achieved by applying $1\times 1$ convolutions to aggregate the pixel-wise cross-channel context followed by $3\times 3$ depth-wise convolutions to encode channel-wise spatial context, yielding $Q=W^{Q}_{d}W^{Q}_{p}$, $K=W^{K}_{d}W^{K}_{p}$, where $W^{(\cdot)}_{p}$ is the $1\times 1$ point-wise convolution and $W^{(\cdot)}_{d}$ is the $3\times 3$ depth-wise convolution. We use bias-free convolutional layers in the network. Whereas the value $(V)$, different from existing works, contains four blocks, each containing a $1 \times 1$ convolution to reduce the channel dimension. The first block just reduces the channel dimension, the second and third blocks contain two dilated temporal convolutions with kernel size 7 and different dilation rates $d_{t}$ to obtain it, and a MaxPool following $1 \times 1$ convolution. The results of the four blocks are concatenated into the output denoted by


\begin{equation}
V= \mathit{TCN}_{d_{2}}(W_{p}Y)\Vert \mathit{TCN}_{d_{3}}(W_{p}Y) \Vert \mathit{Max}(W_{p}Y) \Vert W_{p}Y+Y\text{.}
\end{equation}

Next, we reshape the query and key projections such that their dot-product interaction generates a transposed-attention map $A$ of size $\mathbb{R}^{\widehat{C}\times\widehat{C}}$, instead of the attention map $\mathbb{R}^{\widehat{N}\widehat{T}\times \widehat{N}\widehat{T}}$~\cite{Alexey-17,Ashish-2017}.
Overall, the MHA process is defined as:
\begin{equation}
\mathit{Attention}(\widehat{Q},\widehat{K},\widehat{V})=\widehat{V}\cdot \mathrm{softmax}(\widehat{K}\cdot \widehat{Q}/ \alpha) \text{,}
\end{equation}
where $X$ and $X^{T}$ are the input and output feature maps; $\widehat{Q}, \widehat{V} \in \mathbb{R}^{\widehat{N}\widehat{T} \times \widehat{C} }$, and $\widehat{K}\in \mathbb{R}^{\widehat{C}  \times \widehat{N}\widehat{T}}$ are matrices obtained after reshaping tensors from the original size $\mathbb{R}^{\widehat{N}\times \widehat{T} \times \widehat{C}}$. Here, $\alpha$ is a learnable scaling parameter to the magnitude of the dot product of $\widehat{K}$ and $\widehat{Q}$ before the softmax function is applied.
Similar to the conventional multi-head SA~\cite{Alexey-17}, we divide the number of channels into heads and learn separate attention maps in parallel.
After that, we use GDFN~\cite{Syed-2022} of two fundamental modifications gating mechanism, and depthwise convolutions to improve representation learning.

Finally, we use the average pooling to joint-level temporal fusion features ${\widehat{P}_{1}, ..., \widehat{P}_{V} |\widehat{P}_{i} \in \mathbb{R}^{C\times T}}$ to obtain the skeleton-level feature $\widehat{P'}$.
The skeleton-level temporal fusion feature takes the skeleton-level stride temporal features from the feature extractor.
Joint-level temporal fusion is then applied to merge $\widehat{P'}$ into each $X'_{i}$.
Like DG-TCN, each instance of the skeleton-level temporal fusion feature contains a learned parameter $\varphi \in \mathbb{R^{V}} $.
After the joint-level adaptive element-wise addition with skeleton-level temporal fusion, the feature for joint $i$ is $\widehat{X}_{i} +\varphi_{i}\widehat{P} $, which will be further processed with a $1 \times 1$ convolution to get the output.

\section{Experiments}

\subsection{Dataset}
\textbf{NTU RGB+D} \cite{Shahroudy_2016_CVPR} is a popular resource for recognizing human actions based on skeletons. It consists of 56,880 sequences of such actions.
There are two evaluation benchmarks: Cross-Subject (X-Sub) and Cross-View (X-View).
The training and test sets are drawn from two non-overlapping sets of 20 subjects each in X-Sub. In X-View, the training set is made up of 37,920 samples captured by cameras 2 and 3, while the test set includes 18,960 sequences captured by camera 1. \\
\textbf{NTU RGB+D 120} \cite{Liu2019-on} is an extended version of the NTU RGB+D dataset, containing an additional 57,367 skeleton sequences across 60 additional action classes, for a total of 120 action classes. The authors propose two benchmark evaluations: Cross-Subject which in NTU RGB+D, requires differentiation between two groups of subjects, and each group consists of 53 volunteers and Cross-Setup (X-Setup) which data are acquired in different configurations.

\begin{table}[]
\caption{Partition strategies on NTU RGB+D 120(X-sub).}
\label{table:data_type1}
\centering
\small
\begin{tabular}{ccc}
\hline
Partition Strategy                              & FLOPS & Acc(\%)       \\ \hline
DSTA \cite{dstanet_accv2020}                    & 64.7G & 86.6          \\
EfficientGCN-B4 \cite{song2022constructing}     & 15.2G & 88.3          \\
Hyperformer (Joint Only) \cite{Yuxuan_2023}     & 14.8G & \textbf{86.6} \\
Hands                                           & 4.1G  & 86.3          \\
Legs                                            & 4.1G  & 86.0          \\
Upper, Lower                                    & 4.3G  & 86.4          \\
Wrist, Ankle                                    & 4.7G  & 86.5          \\
Hands Legs, Upper, Lower, Wrist Ankle           & 18.3G & \textbf{89.6} \\ \hline
\end{tabular}
\label{table-body-part}
\end{table}


\subsection{Ablation Study}

\begin{figure}[h]
  \def\@captype{table}
  \begin{minipage}[]{.4\textwidth}
    \caption{Effect of LST on Different Skeleton Encoders}
    \small
\begin{tabular}{lcc}
\hline
\multicolumn{1}{c}{\multirow{2}{*}{Backbone}} & \multicolumn{2}{c}{Acc (\%)}                                          \\
\multicolumn{1}{c}{}      & \multicolumn{1}{c}{w./o.} & \multicolumn{1}{c}{w.} \\ \hline
\hline
ST-GCN                    & 82.6            & \textbf{84.6 (↑2.0)}         \\
CTR-baseline              & 83.7            & \textbf{85.5 (↑1.8)}         \\
CTR-GCN (single scale)    & 84.6            & \textbf{86.0 (↑1.4)}         \\
CTR-GCN (multi scale)     & 84.9            & \textbf{86.0 (↑1.1)}         \\
LST                       & 85.5            & \textbf{85.9 (↑0.4)}         \\  \hline
\end{tabular}
    \label{table-backbone}
  \end{minipage}
  \hfill
  \begin{minipage}[]{.43\textwidth}

    \includegraphics[width=\textwidth]{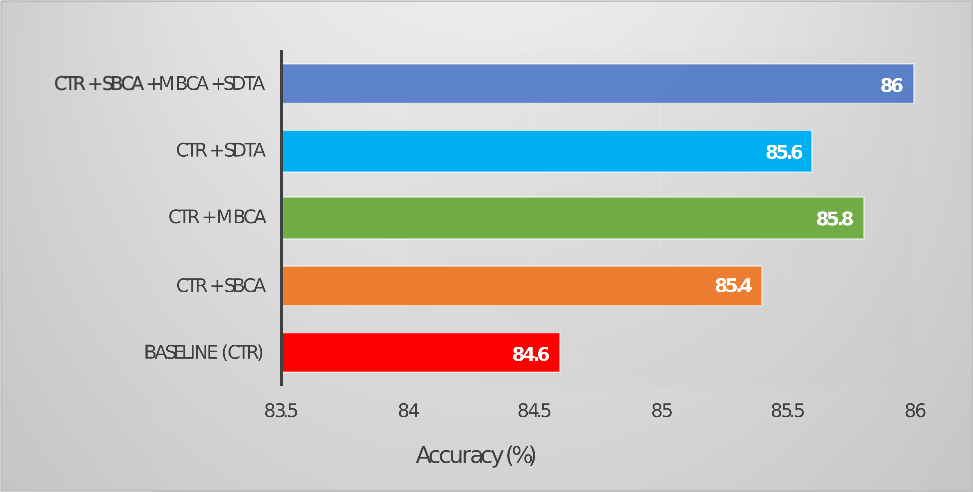}

    \caption{Impact of integrating our contributions in the baseline}
    \label{fig-impart}
  \end{minipage}
\end{figure}

In this section, we evaluate the performance of our Spatial Temporal Effective body-Part Cross Attention Transformer on the X-sub benchmark of the NTU RGB+D 120 dataset. 
\subsubsection*{Partition Strategies}
We test each body-part cross-attention partition strategy for STEP-CATFormer and the results are shown in Table \ref{table-body-part}. Hands, Legs, Wrist, and Ankle represents each hand, leg, wrist, and ankle joint with the remaining other joints cross-attention. And Upper and Lower represents upper joints with lower joints cross-attention. Finally, using more parts and multi-part cross-attention could steadily increase the performance, and it saturates at $ 86.0$\% when using 6 parts cross-attention, and it improves over the baseline by $1.4$\% in X-sub joints. And in 4 ensembles strategy (Joint, Joint-Motion, Bone, Bone-Motion) it saturates at $ 89.6$\% when using 6 parts cross-attention improves over the baseline by $0.7$\%.

\subsubsection*{Impact of the proposed contributions}
This analysis evaluates the impact of our spatiotemporal enrichment module and query-class classifier. Our Spatiotemporal Cross-Attention module includes sub-modules SBCA, MBCA, and SDTA. Figure \ref{fig-impart} compares the performance of our three contributions (Spatiotemporal Cross-Attention module and query-class classifier) with the baseline CTR-GCN. The baseline CTR-GCN has an action classification accuracy of $84.6\%$ (represented by the \textcolor{red}{red bar}). Integrating SBCA, which enriches the spatial context of single body-part cross-attention features before temporal modeling, results in $85.4\%$ accuracy (represented by the \textcolor{orange}{orange bar}). Similarly, the integration of MBCA, which enriches the spatial context of multi-body-part cross-attention features, results in a $1.2\%$ gain (represented by the \textcolor{green}{green bar}). The integration of SDTA, which enriches the temporal context of frame-level features, further improves the accuracy to $85.6\%$ (represented by the \textcolor{cyan}{light blue bar}). Finally, integrating the query-class classifier further enhances feature discriminability, resulting in an accuracy of $86.0\%$ (represented by the \textcolor{blue}{blue bar}). The final STEP-CATFormer framework achieves a $1.4\%$ improvement over the baseline CTR-GCN (\textcolor{red}{red bar}). Our proposed STEP-CATFormer is decoupled from the network architecture and could be used to improve various skeleton encoders. Table \ref{table-backbone} shows the experimental results of applying STEP-CATFormer to ST-GCN, CTR-baseline, CTR-GCN, and LST. STEP-CATFormer brings consistent improvements ($0.4$--$2.0\%$) over the original models with no additional computational cost at inference, demonstrating the effectiveness and generalizability of STEP-CATFormer.

\subsection{Comparison with the State-of-the-art}

\begin{table*}[tb]
\centering
\caption{Performance on the NTU RGB+D and NTU RGB+D 120 dataset. 
}
\small
\begin{tabular}{lccccc}
\hline
\multicolumn{1}{c}{\multirow{2}{*}{Methods}}        & \multicolumn{2}{c}{NTU-60}    & \multicolumn{2}{c}{NTU-120}                                     \\
\multicolumn{1}{c}{}                                & \multicolumn{1}{l}{X-Sub (\%)} & \multicolumn{1}{l}{X-View (\%)} & \multicolumn{1}{l}{X-Sub (\%)} & \multicolumn{1}{l}{X-Set (\%)} \\ \hline
\multicolumn{5}{l}{GCN-based Methods}                                                                                                                                                                                                                                          \\ \hline
ST-GCN\cite{Yan_2018_AAAI}       & 81.5      & 88.3    & 70.7    & 73.2    \\
CTR-GCN\cite{Chen2021-lx}        & 92.4      & 96.8    & 88.9    & 90.6    \\
DG-STGCN\cite{Haodong-2022}      & \textbf{93.2}      & \textbf{97.5}     & 89.6    & \textbf{91.3}       \\
LST\cite{Wangmeng-2022}          & 92.9      & 97      & \textbf{89.9}    & 91.1    \\
Info-GCN\cite{Chi_2022_CVPR}     & 93.0      & 97.1    & 89.8    & 91.2             \\ \hline
\multicolumn{5}{l}{Transformer-based Methods}                                                                                                                                                                                   \\ \hline
ST-TR\cite{Plizzari2020-lc}        & 89.9      & 96.1     & 82.7   & 84.7    \\
STST\cite{Zhang2021-acm}           & 91.9      & 96.8     & -      & -       \\
IIP-Transformer\cite{Wang2021-ua}  & 92.3      & 96.4     & 88.4   & 89.7    \\
FG-STFormer\cite{Gao2022-pd}       & 92.6      & 96.7     & 89.0   & 90.6    \\
Hyperformer\cite{Yuxuan_2023}      & 92.6      & 96.5     & 89.9   & 91.2     \\ \hline


\begin{tabular}{@{}l@{}}STEP CATFormer\\ (CTR-GCN Feature Extractor)\end{tabular}& 93.0    & 96.9    & 89.6    & 90.8       \\\hline

\begin{tabular}{@{}l@{}}STEP CATFormer\\ (LST Feature Extractor)\end{tabular} & \textbf{93.2}    & \textbf{97.3}    & \textbf{90.0}    & \textbf{91.2}        \\

\end{tabular}
\label{table-stepcat}
\end{table*}

We compare our method with the state-of-the-art (SOTA) in Table \ref{table-stepcat}.
For a fair comparison, we use the 4 ensembles strategy (Joint, Joint-Motion, Bone, Bone-Motion) as it is adopted by most of the previous methods.
We also show the ensemble results as well as the individual results.
The results show that our proposed STEP-CATFormer method consistently high-performance one the state-of-the-art.
Our method outperforms all existing transformer-based methods under nearly all evaluation benchmarks on NTU-60 and NTU-120, including the latest method Hyperformer \cite{Yuxuan_2023}.
Besides, our method outperforms CTR-GCN by 0.8\% on NTU-60 X-Sub, and 0.2\% both on NTU-60 X-Sub and X-View compared to Info-GCN \cite{Chi_2022_CVPR}.
It also outperforms LST \cite{Wangmeng-2022} by 0.3\% both on them.
Table \ref{table-stepcat} shows that STEP-CATFormer outperforms CTR-GCN on the largest dataset, NTU RGB+D 120, by a significant margin of 1.1\% on cross-subject and 0.6\% on cross-set.
Although Info-GCN also performs well in 6 ensemble strategies on this dataset, STEP-CATFormer still higher with 0.1\% and 0.1\% in 4 ensemble strategies, respectively.

\section{Discussion and Conclusion}
Our research paper is dedicated to the optimization of STEP-CATFormer to improve skeleton-based action recognition using Kinect camera, with a particular focus on the caracteristics  of two specific skeleton datasets.
While our proposed model may not necessarily be the most effective for other tasks, our approach of taking inherent relationships into account has the potential to enhance the use of Transformers in various applications.

The paper is focused on improve recognizes the action performed on the Kinect sensor real-time skeletal tracking to understands what the user is doing, then responding exactly what they need to improve the quality of the game play.

We present a novel spatio-temporal effective body-part cross-attention transformer network (STEP-CATFormer) for skeleton-based action recognition.
In the spatial dimension, it learns joint and body-part correlations for adaptively sampled joint body-part relationships, which captures the discriminative and comprehensive spatial dependencies.
In the temporal dimension, it explicitly learns dynamic temporal relations with a dilation convolution transformer, enabling the network to capture rich motion patterns effectively.
STEP-CATFormer high-performed on state-of-the-art on NTU RGB+D, and NTU RGB+D 120 benchmarks.

\bibliography{egbib}
\end{document}